\documentclass[letterpaper]{article} 
\usepackage{aaai23}  
\usepackage{times}  
\usepackage{helvet}  
\usepackage{courier}  
\usepackage[hyphens]{url}  
\usepackage{graphicx} 
\usepackage{amsmath,amssymb,amsfonts}
\usepackage{algorithmic}
\usepackage{graphicx}
\usepackage{textcomp}
\usepackage{xcolor}
\usepackage{booktabs}
\usepackage{subfig}
\def\UrlFont{\rm}  
\usepackage{natbib}  
\usepackage{caption} 
\usepackage{algorithm}
\usepackage{algorithmic}

\usepackage{newfloat}
\usepackage{listings}
\DeclareCaptionStyle{ruled}{labelfont=normalfont,labelsep=colon,strut=off} 
\floatstyle{ruled}
\newfloat{listing}{tb}{lst}{}
\floatname{listing}{Listing}
\nocopyright

\title{LB-SimTSC: An Efficient Similarity-Aware Graph Neural Network for Semi-Supervised Time Series Classification}

\author {
    Wenjie Xi, \textsuperscript{\rm 1}\footnote{These authors contributed equally.}
    Arnav Jain, \textsuperscript{\rm 2}\footnotemark[1]
    Li Zhang, \textsuperscript{\rm 1}
    Jessica Lin \textsuperscript{\rm 1}
}
\affiliations {
    \textsuperscript{\rm 1} George Mason University, Fairfax, Virginia, USA\\
    \textsuperscript{\rm 2} Thomas Jefferson High School for Science and Technology, Alexandria, Virginia, USA\\
    wxi@gmu.com, arnavj22@gmail.com, lzhang18@gmu.edu, jessica@gmu.edu
}

\begin{document}

\maketitle

\begin{abstract}
Time series classification is an important data mining task that has received a lot of interest in the past two decades. Due to the label scarcity in practice, semi-supervised time series classification with only a few labeled samples has become popular. Recently, Similarity-aware Time Series Classification (SimTSC) is proposed to address this problem  by using a graph neural network classification model on the graph generated from pairwise Dynamic Time Warping (DTW) distance of batch data. It shows excellent accuracy and outperforms state-of-the-art deep learning models in several few-label settings. However, since SimTSC relies on pairwise DTW distances, the quadratic complexity of DTW limits its usability to only reasonably sized datasets. To address this challenge, we propose a new efficient semi-supervised time series classification technique, LB-SimTSC, with a new graph construction module. Instead of using DTW, we propose to utilize a lower bound of DTW, LB\_Keogh, to approximate the dissimilarity between instances in linear time, while retaining the relative proximity relationships one would have obtained via computing DTW. We construct the pairwise distance matrix using LB\_Keogh and build a graph for the graph neural network. We apply this approach to the ten largest datasets from the well-known UCR time series classification archive. The results demonstrate that this approach can be up to 104x faster than SimTSC when constructing the graph on large datasets without significantly decreasing classification accuracy.
\end{abstract}

\section{Introduction}\label{intro}
\begin{figure*}[t]
\begin{center}
\vspace{-2mm}
\centerline{\includegraphics[width=0.82\textwidth,
height=0.28\textwidth]{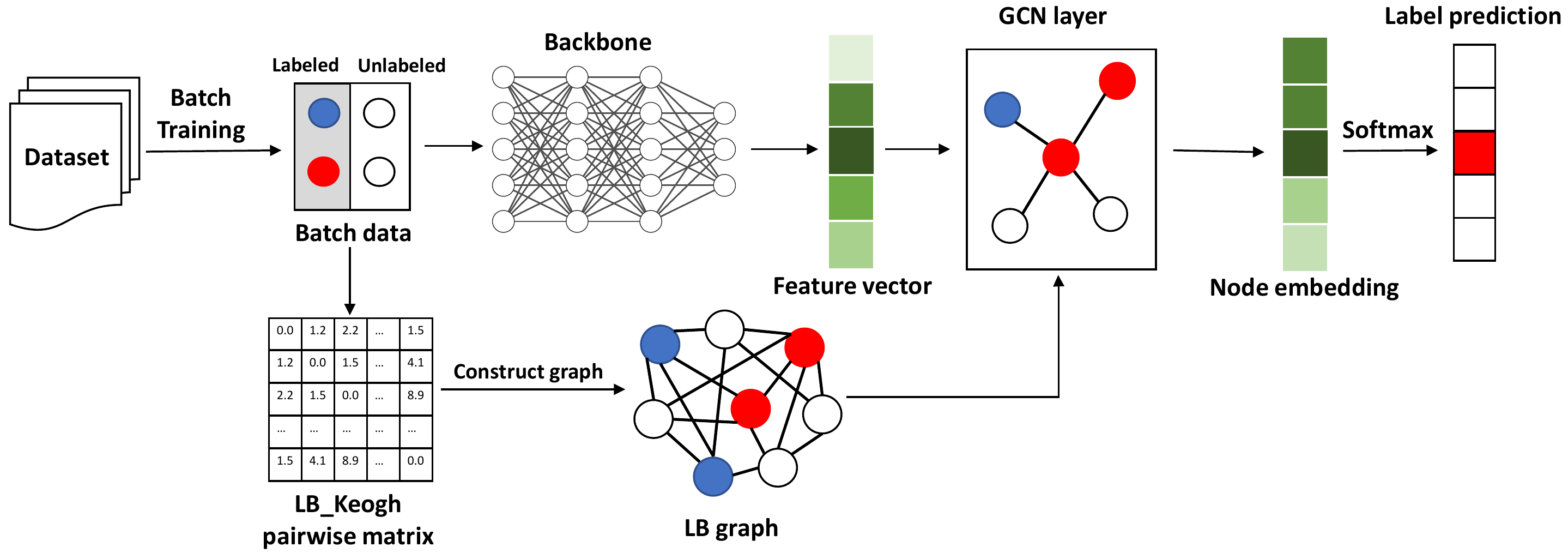}}
\caption{Overall framework of LB-SimTSC.}
\label{overview}
\end{center}
\vspace{-8mm}
\end{figure*}
{\color{black} 
Time series classification is an important research area and attracts a great amount of attention~\cite{zhang2020tapnet}. In practice, due to the difficulty and high cost of obtaining labeled data, there might only be a few labeled instances per class for training. This becomes one of the greatest challenges in time series classification, and motivates researchers to investigate semi-supervised solutions~\cite{wei2006semi}. Recently, Zha et al.~\cite{zha2022towards} proposed SimTSC, which combines batch Graph Convolution Network (GCN)~\cite{kipf2016semi} with Dynamic Time Warping (DTW), a commonly used 
distance measure that allows non-linear alignments in time. DTW has been shown to be more robust than other distance measures like Euclidean Distance in tasks such as classification and similarity search~\cite{bagnall2017great}. Specifically, SimTSC builds a graph from the DTW distance matrix; each node corresponds to a time series instance to be classified, and the edge is generated through DTW distance. SimTSC is versatile for three reasons. First, it only requires a batch dynamic subgraph and does not need the complete graph for the entire data to be stored in the GPU for training. Second, compared with the classic graph generation process, which simply uses the k-nearest neighbors (k-NN) to generate a static graph, SimTSC uses a graph via local k-NN generated from batch data, which allows the information in labeled data to be efficiently passed to the unlabeled data.  
Third, SimTSC is able to learn a warping-aware representation by inheriting the advantages of using DTW~\cite{sakoe1971dynamic}. 

Despite the advantages, SimTSC is not scalable for large data. To construct the graph, SimTSC requires the full  pairwise computation of DTW distances, a process with a high computational cost of $O(L^2)$, where L is the time series instance length. For example, for the famous \textit{StarLightCurves} dataset, which contains 9,236 time series of length 1,024, DTW takes around two days to calculate the distances between all pairs. Furthermore, since DTW computation uses dynamic programming, it is not parallelizable with GPU computations. While several techniques have been proposed to speed up DTW computation, they cannot be applied in this case since they work by pruning unnecessary computations for the similarity search problem, whereas SimTSC requires the full pairwise distance matrix for training. 

To address this problem, we propose a new efficient graph based time series classification technique, \textbf{L}ower \textbf{B}ound \textbf{Sim}ilarity-aware \textbf{T}ime \textbf{S}eries \textbf{C}lassifier (LB-SimTSC) with a new efficient graph construction approach. Instead of computing the pairwise distance matrix using DTW, we propose to use a lower-bounding distance for DTW, which is much cheaper to compute. In particular, we adapt LB\_Keogh, a popular lower bound for DTW~\cite{keogh2005exact}, with $O(L)$ time complexity. For the \textit{StarlightCurves} dataset, it would only take about 30 minutes to compute all pairwise distances using LB\_Keogh instead of 2 days. The idea of lower bound is originally proposed to speed up the 1-NN similarity search problem by providing a reasonable, efficient-to-compute approximation of the actual distance and allowing us to prune some candidates from consideration. It is not designed as a similarity measure by itself. However, we observe that LB\_Keogh matrix can be used as a replacement for DTW to provide the guidance needed by GCN and help GCN learn a similar warping-aware representation. We demonstrate through empirical evaluation that using the approximation matrix can achieve similar performance as the original SimTSC, with reduced running time by two orders of magnitude.

In summary, our contribution is as follows: 
\begin{itemize}
    \item In this paper, we investigate an important and challenging problem of semi-supervised time series classification with only a few labeled samples. 
    \item We propose a new efficient graph-based time series classification technique, LB-SimTSC, with a new graph construction module that takes only linear time.  
    \item Our experimental results show that our proposed LB-SimTSC achieves similar performance with SimTSC, but with much faster graph construction for large data. 
    \item Our work demonstrates that it is possible to leverage a much more efficient lower bound distance-based graph to guide GCN, while still maintaining the same ability as using the more expensive DTW distance. 
\end{itemize}

\section{Related Work} 
\noindent\textbf{Graph-based Time Series Classification} Most existing graph-based time series classification methods either require actual graphs (e.g., transportation) as an input or do not consider semi-supervised settings with few labels in each class~\cite{, alfke2021empirical}. The most closely related work is SimTSC~\cite{zha2022towards}. SimTSC constructs a graph from the DTW pairwise matrix, where each node represents a time series, and the connection between nodes represents their similarity. The computation of the matrix is very time-consuming for large data. 
Our proposed method LB-SimTSC takes an alternative approach to address the efficiency bottleneck by utilizing LB\_Keogh, a lower bound of DTW, to construct the graph. LB-SimTSC significantly reduces the graph construction time from quadratic to linear in the length of instances. It is thus more scalable for large datasets while maintaining comparable performance as SimTSC. 

\noindent\textbf{Lower Bound of DTW} Lower bound of DTW is introduced to speed up time series similarity search. 
Since lower bounding measures are often much cheaper to compute than DTW, they are often used to approximate the minimum distance between a pair of sequences. Such approximation allows the quick elimination of certain unqualified candidates without computing the actual distances in similarity search. 
Driven by the need for efficient DTW-based similarity search, many lower bounding distance measures such as LB\_Yi, LB\_Kim, and LB\_Keogh~\cite{yi1998efficient,kim2001index, keogh2005exact}
have been proposed. Among them, LB\_Keogh~\cite{keogh2005exact} is simple to compute and remains one of the most competitive and tightest lower bounds for DTW-based query tasks. 

To the best of our knowledge, we are the first to explore the direction of using a lower bounding measure for graph construction for graph-based semi-supervised time series classification models.
}
\section{Problem Formulation}\label{preliminaries} 

In this paper, we study a time series semi-supervised classification problem with the transductive setting~\cite{zha2022towards}, where test data without labels can be seen during training time. Consider a training data $\mathcal{X}^{train}$ consisting of n time series instances $X_i$ of length $L$, with only a small $n'$ instances $\mathcal{X}^{labeled} = [X_1, X_2, \cdots, X_{n'}] $ with known labels $Y^{labeled} = [y_1, y_2, \cdots, y_{n'}]$ where $y_i \in \{1, \cdots, C\}$ classes, and $\mathcal{X}^{unlabeled}$ data without labels. During training, our proposed model is trained on $\mathcal{X}^{labeled}$ and $Y^{labeled}$ with $\mathcal{X}^{unlabeled}$. Our goal is to train a model that correctly predicts the labels of test data $\mathcal{X}^{test}$. 

\section{Methodology}\label{method}

Fig.~\ref{overview} shows the overall framework of the proposed method. The framework consists of three steps: \textbf{1) Batch Sampling}. Given a dataset $(\mathcal{X}, \boldsymbol{Y})$, we first form a batch consisting of an equal number of labeled and unlabeled data $\mathcal{X}^{B}$. \textbf{2) LB\_Keogh Lower Bound Graph Construction}. We compute a pairwise lower bound distance matrix with respect to $\mathcal{X}^{B}$ to generate a batch LB-graph $G^{B}$ that describes the approximate warping-aware (dis)similarity between pairs of instances within each batch. The feature vector is obtained from the backbone network. \textbf{3) Graph Convolution and Classification} The feature vector is passed through Graph Convolution Network (GCN)~\cite{kipf2016semi} to aggregate the features across the generated $G^{B}$. Then the obtained node embedding is passed through a fully connected neural network and performs the classification task. 

There are two main advantages of the framework: We can pass the label information from labeled instances to unlabeled instances through batched data. In addition, the GCN aggregates the information of similar instances under warping captured by the generated batch LB-graph $G^B$ and achieves the preservation of warping-aware similarity on the instance level in the latent space. 
\subsubsection{Batch Sampling}
To form a batch $\mathcal{X}^B$ of size $m$, we sample $m/2$ number of instances from labeled data $(\mathcal{X}^{labeled}, Y^{labeled})$, and $m/2$ number of instances from unlabeled data $\mathcal{X}^{unlabeled}$. In this step, we ensure that each class has equal samples to avoid imbalance in the learning process. The batch data will be used to compute the embedding and the proposed batch LB-Graph. 

\subsubsection{LB\_Keogh Graph Construction}
We introduce our proposed LB\_Keogh Graph (LB-Graph) construction in this subsection. 

LB\_Keogh ~\cite{keogh2005exact} is originally proposed as a lower bound to prune unnecessary computations of DTW when performing similarity search. It is much cheaper to compute than the actual DTW distance, but still maintains the ability to allow warping when comparing two time series. 

Given two time series instances $X_i$ and $X_j $ from batch $\mathcal{X}^{B}$ , LB\_Keogh creates an envelope consisting of one upper bound and one lower bound time series. As shown in Eq.~\ref{eqn: UL}, for each timestamp $k$: 
\begin{equation}
\label{eqn: UL}
\begin{aligned}
    u_{i, k} &= \max([X_{i, k-r}, X_{i, k-r+1},\cdots, X_{i, k+r}])\\
    l_{i, k} &= \min([X_{i, k-r}, X_{i, k-r+1},\cdots, X_{i, k+r}])
\end{aligned}
\end{equation}
where $r$ is the allowed range of warping. Then the LB\_Keogh distance $d^{LB}$ between instance $X_i$ and $X_j$ is calculated by Eq. \ref{eq2}:
\vspace{-2mm}

\begin{equation}
    d^{LB}(X_i, X_j) = \sqrt{\sum_{k=1}^{L}\left\{\begin{array}{l}
    (X_{j, k} - u_{i, k})^2, \quad if X_{j, k} > u_{i, k} \\
    (X_{j, k}-l_{i, k})^2, \quad if X_{j, k} < l_{i, k} \\
    0, \quad otherwise\\
\end{array}\right.}
\label{eq2}
\end{equation}
\vspace{-4mm}

\begin{figure}[t]
\centerline{\includegraphics[width=0.32\textwidth,height=0.15\textwidth]{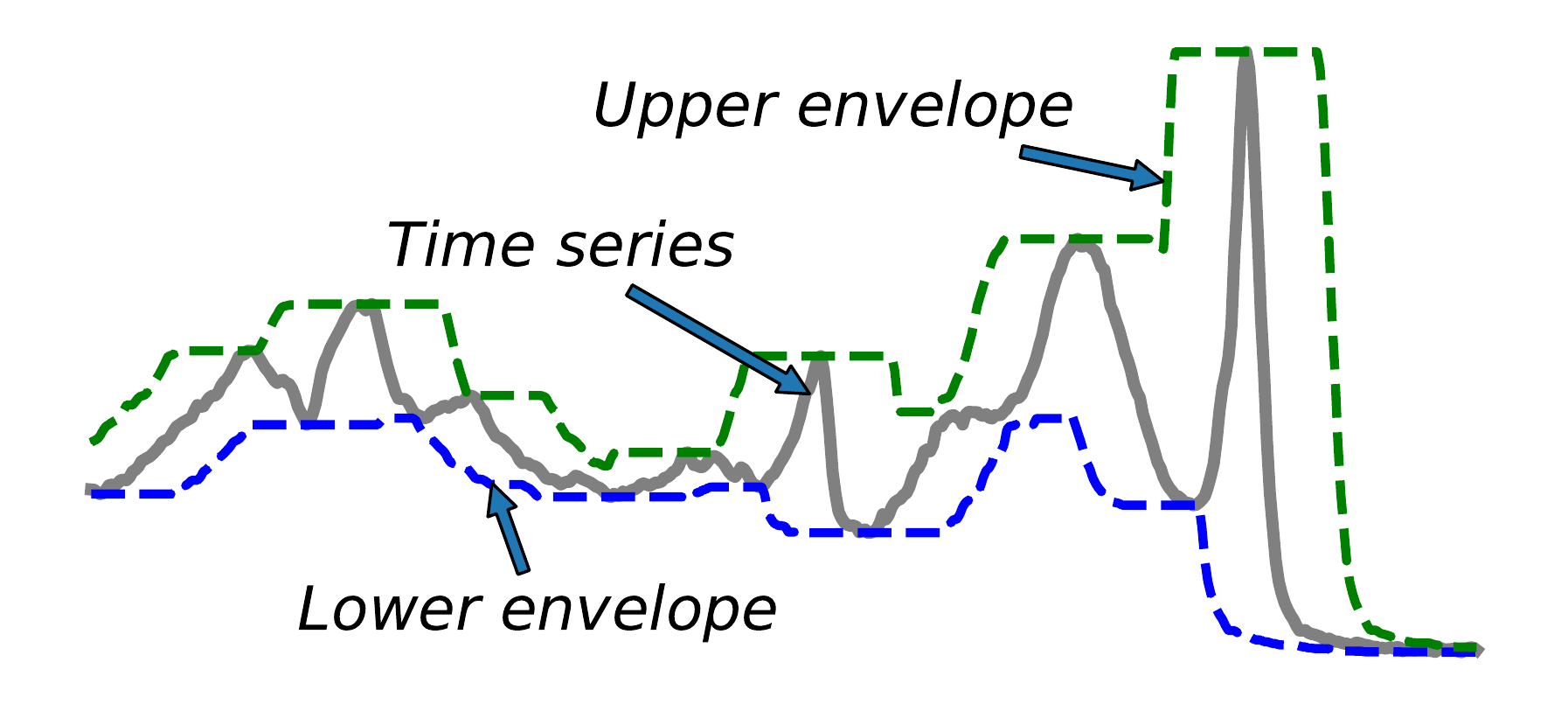}}

\caption{A visual intuition of the lower envelope (blue dotted line) and upper envelope (green dotted line) of a sample time series (gray solid line) generated by LB\_Keogh.}
\label{fig2}
\end{figure}

The pairwise LB\_Keogh matrix $D_{LB}$ given a batch $\mathcal{X}^B$ is hence defined as 
\begin{equation}
    \label{eqn:LB_matrix}
    D_{LB}(i, j) = d^{LB} (X_i, X_j), \quad X_i\in \mathcal{X}^B, X_j \in \mathcal{X}^B. 
\end{equation}

\subsubsection{Graph Construction from $D_{LB}$} 
We next generate the graph $G^B(V, E)$ from $D_{LB}$. We perform a two-step approach to compute the edges. 

We first convert the LB\_Keogh distance to similarity by: 
\begin{equation}
    \label{eqn:A}
    A_{i, j} = \frac{1}{\exp{(\alpha D_{i, j})}}.
\end{equation}
where $\alpha$ is scaling parameter. 

Then we generate a sparse graph based on $A$. For each row $A_i=[A_{i,1},\dots,A_{i,m}]$ in $A$, we select $K$ candidates to form the sparse adjacency matrix. Specifically, to form a vector $Q_i$, we select random $K$ candidates with zero LB\_Keogh distance, which indicates that the shapes of these instances fall within the enveloped area (i.e. they are likely to be similar to $X_i$). If there are fewer than $K$ candidates with zero distance, then we pick $K$ instances with the smallest LB\_Keogh distances.

To incorporate this idea, we compute a sparse adjacency matrix $\bar{A}_{i, j}$ by:  
\begin{equation}
    \bar{A}_{i, j} = \left\{\begin{array}{l}
    1/K, \quad j \sim \mathcal{Q}_i \quad\&\quad |\mathcal{Q}_i|\geq K; \\
    A_{i, j}, \quad A_{i, j}\in \arg \text{TopK}(A_{i})\quad \&\quad |\mathcal{Q}_i|< K; \\
    0, \quad\quad \text{otherwise}. \\
\end{array}\right.
\label{eqn:a_bar}
\end{equation}
where $\mathcal{Q}_i=\{k|D_{i,k}=0\}$, and we randomly pick $K$ samples from $\mathcal{Q}_i$. 

Now we are ready to build the batch graph $G^{B}(V, E)$ where $|V| = m$, and the weights between node $i$ and node $j$ is defined as: 
\begin{equation}
    E_{i, j} = \frac{\bar{A}_{i,j}}{\sum_{b = 1}^m \bar{A}_{i,b}}.  
    \label{eqn:e}
\end{equation}

The overall LB graph construction algorithm is summarized in Algorithm 1.

\begin{algorithm}[]
\caption{LB-Graph Construction}\label{graph construction}
\begin{algorithmic}[1] 
\STATE \textbf{Input}: $\mathcal{X}^{B}$: batch data, $r$: warping range, $\alpha$: scaling factor, $K$: number of top neighbors \\
\STATE \textbf{Output}: $G^{B}$: the graph of the current batch
\\{\color{blue} /* Compute $D_{LB}$ via Eqn.~\ref{eqn:LB_matrix} */}
\STATE $D$=ComputePairwiseLB($X_{batch}$)
\\{\color{blue} /* Compute $A$ via Eqn.~\ref{eqn:A} */}
\FOR{each $D_{i, j}$ in $D$} 
\STATE $A_{i,j}= 1/\exp(\alpha D_{i,j})$\\
\ENDFOR
\\{\color{blue} /* Compute $G^{B}$ via Eqn.~\ref{eqn:a_bar} and Eqn.~\ref{eqn:e}. */}
\STATE $G^{B} =$ GenerateGraph$(A, K)$ \\
\STATE \textbf{return} $G^{B}$
\end{algorithmic}
\end{algorithm}
\vspace{-3mm}

\subsubsection{Graph Convolution and Classification}
Given a backbone network $f(.)$, we compute the feature embedding network $z = f(\mathcal{X}^B)$. Then the embedding $z$ is passed through a series of graph convolutional network (GCN) layers and used to perform classification. Particularly, 
\begin{align*}
    z^{l} &= GCN(G^{B}, z^{l-1}) \\
    o &= Softmax(z^l).
\end{align*}
where $l$ is the number of layers in this case. In our work, we use the same backbone network as SimTSC~\cite{zha2022towards}, which is a ResNet~\cite{he2016deep} with three residual blocks. 

\subsection{Advantages of LB-SimTSC}
Compared with SimTSC~\cite{zha2022towards}, our proposed method has two advantages. First, we construct batch LB\_Keogh graph, which can be seen as an alternative but efficient solution to measure warping similarity. Compared to the original DTW distance computation, which is expensive via dynamic programming or recursion, LB\_Keogh is simple and only requires $O(L)$ time complexity. Second, from Eqn 2, we can see that $D_{LB}$ is highly parallelizable, thus more suitable to implement in GPU. 

Moreover, GCN does not require exact weights to perform aggregate information to propagate label information. Therefore, although LB\_Keogh matrix $D_{LB}$ containing zero values provides a less accurate warping measure than DTW, all of these instances fall into the envelop area and share some degree of warping similarity. Thus, a random subset of zero value in $D_{LB}$ is sufficient to guide the training and propagate labels. In our experiments, we find that the empirical performance of LB-SimTSC is statistically comparable with the performance of SimTSC. 
\section{Experimental Evaluation}
In this section, we evaluate both efficiency and performance of our proposed LB-SimTSC. 
\subsubsection{Datasets} 
Our evaluation will focus on demonstrating the efficiency of LB-SimTSC, as well as the ability to achieve competitive performance even with few labels available. We choose all large datasets from the UCR classification archive~\cite{dau2019ucr}. Specifically, we choose datasets with sequence lengths greater than 500, and the number of instances greater than 2000. A total of ten datasets are included. 
We follow the same dataset generation protocol as SimTSC, whereas we first split every dataset into training set (80\%) and test set dataset (20\%). We randomly sample $ \beta$ instances for each class from the training data to form $\mathcal{X}^{labeled}$ and $Y^{labeled}$. The remaining training data is used as $\mathcal{X}^{unlabeled}$. We test six different settings $\beta = \{5,10,15,20,25,30\}$. We report the classification accuracy for each case. 
\subsubsection{Hyperparameters} We keep all the settings of the baselines, consistent with the original paper. For our model, we fix the warping range $r$ to be 5\% of time series length, and empirically set the scaling factor $\alpha$ to 11. All other settings are consistent with SimTSC. 
\begin{table}[htbp]
\caption{Accuracy scores of LB-SimTSC (ours) and 1NN-DTW on ten large datasets. The p-value is calculated using the Wilcoxon signed-rank test (one-sided).}
\scalebox{0.58}{
\begin{tabular}{@{}lrrrrrr@{}}
\toprule
              & \multicolumn{1}{l}{5 labels} & \multicolumn{1}{l}{} & \multicolumn{1}{l}{10 labels} & \multicolumn{1}{l}{} & \multicolumn{1}{l}{15 labels} & \multicolumn{1}{l}{} \\ \midrule
Datasets      & 1NN-DTW                      & LB-SimTSC            & 1NN-DTW                       & LB-SimTSC            & 1NN-DTW                       & LB-SimTSC            \\ \midrule
FordA         & 0.540                        & \textbf{0.793}       & 0.531                         & \textbf{0.826}       & 0.545                         & \textbf{0.816}       \\
FordB         & 0.627                        & \textbf{0.812}       & 0.603                         & \textbf{0.806}       & 0.628                         & \textbf{0.807}       \\
NIFECGT1 & \textbf{0.665}               & 0.662                & 0.704                         & \textbf{0.743}       & 0.710                         & \textbf{0.830}       \\
NIFECGT2 & \textbf{0.757}               & 0.689                & 0.788                         & \textbf{0.798}       & 0.817                         & \textbf{0.856}       \\
UWGLAll       & \textbf{0.779}               & 0.447                & \textbf{0.854}                & 0.549                & \textbf{0.865}                & 0.622                \\
Phoneme       & 0.185                        & \textbf{0.253}       & 0.204                         & \textbf{0.330}       & 0.209                         & \textbf{0.346}       \\
Mallat        & \textbf{0.944}               & 0.908                & 0.960                         & \textbf{0.963}       & \textbf{0.971}                & 0.960                \\
MSST          & 0.784                        & \textbf{0.871}       & 0.796                         & \textbf{0.883}       & 0.834                         & \textbf{0.908}       \\
MSRT          & \textbf{0.790}               & 0.775                & 0.819                         & \textbf{0.891}       & 0.843                         & \textbf{0.916}       \\
SLC           & 0.559                        & \textbf{0.920}       & 0.683                         & \textbf{0.939}       & 0.763                         & \textbf{0.946}       \\ \midrule
wins          & \textbf{5}                   & \textbf{5}           & 1                             & \textbf{9}           & 2                             & \textbf{8}           \\
p-value       & 0.246                        & -                    & 0.042                         & -                    & 0.042                         & -                    \\ \bottomrule
\end{tabular}}
\scalebox{0.58}{
\begin{tabular}{@{}lrrrrrr@{}}
\toprule
              & \multicolumn{1}{l}{20 labels} & \multicolumn{1}{l}{} & \multicolumn{1}{l}{25 labels} & \multicolumn{1}{l}{} & \multicolumn{1}{l}{30 labels} & \multicolumn{1}{l}{} \\ \midrule
Datasets      & 1NN-DTW                       & LB-SimTSC            & 1NN-DTW                       & LB-SimTSC            & 1NN-DTW                       & LB-SimTSC            \\ \midrule
FordA         & 0.535                         & \textbf{0.825}       & 0.548                         & \textbf{0.862}       & 0.550                         & \textbf{0.864}       \\
FordB         & 0.630                         & \textbf{0.820}       & 0.627                         & \textbf{0.836}       & 0.637                         & \textbf{0.847}       \\
NIFECGT1 & 0.720                         & \textbf{0.855}       & 0.732                         & \textbf{0.867}       & 0.729                         & \textbf{0.860}       \\
NIFECGT2 & 0.819                         & \textbf{0.856}       & 0.821                         & \textbf{0.874}       & 0.826                         & \textbf{0.879}       \\
UWGLAll       & \textbf{0.879}                & 0.628                & \textbf{0.886}                & 0.682                & \textbf{0.911}                & 0.677                \\
Phoneme       & 0.258                         & \textbf{0.408}       & 0.277                         & \textbf{0.419}       & 0.282                         & \textbf{0.427}       \\
Mallat        & \textbf{0.971}                & \textbf{0.971}       & 0.971                         & \textbf{0.972}       & \textbf{0.977}                & 0.969                \\
MSST          & 0.838                         & \textbf{0.914}       & 0.846                         & \textbf{0.931}       & 0.853                         & \textbf{0.925}       \\
MSRT          & 0.855                         & \textbf{0.922}       & 0.851                         & \textbf{0.933}       & 0.853                         & \textbf{0.942}       \\
SLC           & 0.798                         & \textbf{0.924}       & 0.807                         & \textbf{0.936}       & 0.780                         & \textbf{0.942}       \\ \midrule
wins          & 2                             & \textbf{9}           & 1                             & \textbf{9}           & 2                             & \textbf{8}           \\
p-value       & 0.043                         & -                    & 0.024                         & -                    & 0.042                         & -                    \\ \bottomrule
\end{tabular}}
\vspace{-2mm}
\label{acc}
\end{table}

\subsubsection{Experiment 1: Comparing with 1NN-DTW}

We compare with the 1-Nearest Neighbor classifier under Dynamic Time Warping Distance (1NN-DTW). 1NN-DTW is a well-known baseline in time series classification~\cite{bagnall2017great} and still performs well even with only few labeled data available~\cite{zha2022towards}. \\

Table~\ref{acc} shows the classification accuracy of 1NN-DTW and our model on the 10 datasets. Our model wins 8 or 9 out of 10 in most cases, and ties with 1NN-DTW when $\beta=5$. To further illustrate the superiority of the accuracy of our model, we perform the one-sided Wilcoxon signed-rank test. The p-values of the Wilcoxon test between two models on all $\beta$ settings are 0.246, 0.042, 0.042, 0.043, 0.043, 0.024 and 0.042, respectively. Among them, except for $\beta = 5$, the rest of the p-values are all less than 0.05. Thus we conclude that our model is significantly better than 1NN-DTW across all settings except 5 labels, which is consistent with the observation in SimTSC. This is potential because the extreme small-sample situation is easier for the classical 1-NN method but more difficult for deep learning to get good performance.  
\subsubsection{Experiment 2: Comparing with SimTSC }
\begin{table}[htbp]
\vspace{-2mm}
\caption{Comparison of graph construction time}
\centering
\scalebox{0.7}{
\begin{tabular}{@{}lrrrrr@{}}
\toprule
Datasets      & Length & Instance & \multicolumn{1}{l}{\begin{tabular}[c]{@{}l@{}}SimTSC\\ (in hours)\end{tabular}} & \multicolumn{1}{l}{\begin{tabular}[c]{@{}l@{}}LB-SimTSC\\ (in hours)\end{tabular}} & Faster \\ \midrule
FordA         & 500    & 4921     & 3.826                                                                           & \textbf{0.127}                                                                              & \textbf{30x}    \\
FordB         & 500    & 4446     & 3.310                                                                           & \textbf{0.099}                                                                              & \textbf{33x}    \\
NIFECGT1 & 750    & 3765     & 4.622                                                                           & \textbf{0.082}                                                                              & \textbf{56x}    \\
NIFECGT2 & 750    & 3765     & 4.678                                                                           & \textbf{0.081}                                                                              & \textbf{58x}    \\
UWGLAll       & 945    & 4478     & 10.385                                                                          & \textbf{0.126}                                                                              & \textbf{82x}    \\
Phoneme       & 1024   & 2110     & 2.841                                                                           & \textbf{0.029}                                                                              & \textbf{96x}    \\
Mallat        & 1024   & 2400     & 3.716                                                                           & \textbf{0.036}                                                                              & \textbf{104x}   \\
MSST          & 1024   & 2525     & 4.014                                                                           & \textbf{0.041}                                                                              & \textbf{96x}    \\
MSRT          & 1024   & 2925     & 5.045                                                                           & \textbf{0.058}                                                                              & \textbf{86x}    \\
SLC           & 1024   & 9236     & 46.240                                                                          & \textbf{0.566}                                                                              & \textbf{82x}    \\ \midrule
Total         &        &          & 88.676                                                                          & \textbf{1.247}                                                                              & \textbf{71x}    \\ \bottomrule
\end{tabular}}
\vspace{-2mm}
\label{running time}
\end{table}

SimTSC~\cite{zha2022towards} is the closest related work. We compare computational time for graph construction, which is the major bottleneck in overall computation cost. For both methods, to ensure a fair comparison, we follow the same implementation process described in Zha et al.~\cite{zha2022towards} via pre-computing the similarity matrix for the entire dataset first, then performing sampling for each batch. 
For our model, LB-SimTSC, we use Python to implement the graph construction process. For SimTSC, since their graph construction code in C is not executable on our computer, we use an equivalent C++ implementation to estimate their running time. We report the actual running time for both methods. 

 Table~\ref{running time} shows the graph construction time in hours for SimTSC and LB-SimTSC on the ten datasets. We can observe that even when we use the slower implementation (Python) for the graph construction module of our model and the much faster implementation (MATLAB) for SimTSC, our model is still 71x faster than SimTSC in total time, and up to 104x faster on one dataset. For the time required to complete all graph construction, our approach reduces it from 3 days 16 hours to only 75 minutes. Note that an increase in time series length or number of instances makes the speed gap of our method more significant than that of SimTSC. The large time saving of our model makes our approach desirable for semi-supervised TSC on large data with few labels.

We also compare the accuracy to demonstrate that LB-SimTSC still achieves competitive performance. To eliminate randomness in deep learning training, we run both methods three times and report the average accuracy. Table~\ref{acc2} shows the classification accuracy of SimTSC and our model. The number of wins of SimTSC is slightly more than ours in all settings. To show the difference in accuracy between the two models, we perform the two-sided Wilcoxon signed-rank test. The p-values between SimTSC and LB-SimTSC are 0.492, 0.375, 0.375, 0.232, 0.695 and 0.492, respectively. All the p-values are less than 0.05, meaning our model does not have a significant difference in accuracy from SimTSC.


\begin{table}[htbp]
\caption{Accuracy scores of LB-SimTSC (ours) and SimTSC on ten large datasets. The p-value is calculated using the Wilcoxon signed-rank test (two-sided).}
\scalebox{0.62}{
\begin{tabular}{@{}lrrrrrr@{}}
\toprule
              & \multicolumn{1}{l}{5 labels} & \multicolumn{1}{l}{} & \multicolumn{1}{l}{10 labels} & \multicolumn{1}{l}{} & \multicolumn{1}{l}{15 labels} & \multicolumn{1}{l}{} \\ \midrule
Datasets      & SimTSC                       & LB-SimTSC            & SimTSC                        & LB-SimTSC            & SimTSC                        & LB-SimTSC            \\ \midrule
FordA         & \textbf{0.795}               & 0.793                & \textbf{0.832}                & 0.826                & \textbf{0.827}                & 0.816                \\
FordB         & 0.768                        & \textbf{0.812}       & 0.800                         & \textbf{0.806}       & 0.802                         & \textbf{0.807}       \\
NIFECGT1 & \textbf{0.692}               & 0.662                & \textbf{0.790}                & 0.743                & \textbf{0.861}                & 0.830                \\
NIFECGT2 & \textbf{0.733}               & 0.689                & \textbf{0.823}                & 0.798                & \textbf{0.866}                & 0.856                \\
UWGLAll       & \textbf{0.458}               & 0.447                & \textbf{0.561}                & 0.549                & 0.601                         & \textbf{0.622}       \\
Phoneme       & \textbf{0.261}               & 0.253                & 0.313                         & \textbf{0.330}       & \textbf{0.372}                & 0.346                \\
Mallat        & \textbf{0.943}               & 0.908                & 0.952                         & \textbf{0.963}       & \textbf{0.968}                & 0.960                \\
MSST          & \textbf{0.880}               & 0.871                & 0.871                         & \textbf{0.883}       & \textbf{0.914}                & 0.908                \\
MSRT          & 0.732                        & \textbf{0.775}       & \textbf{0.907}                & 0.891                & 0.912                         & \textbf{0.916}       \\
SLC           & 0.914                        & \textbf{0.920}       & \textbf{0.948}                & 0.939                & 0.931                         & \textbf{0.946}       \\ \midrule
wins          & \textbf{7}                   & 3                    & \textbf{6}                    & 4                    & \textbf{6}                    & 4                    \\
p-value       & 0.492                        & -                    & 0.375                         & -                    & 0.375                         & -                    \\ \bottomrule
\end{tabular}}
\scalebox{0.62}{
\begin{tabular}{@{}lrrrrrr@{}}
\toprule
              & \multicolumn{1}{l}{20 labels} & \multicolumn{1}{l}{} & \multicolumn{1}{l}{25 labels} & \multicolumn{1}{l}{} & \multicolumn{1}{l}{30 labels} & \multicolumn{1}{l}{} \\ \midrule
Datasets      & SimTSC                        & LB-SimTSC            & SimTSC                        & LB-SimTSC            & SimTSC                        & LB-SimTSC            \\ \midrule
FordA         & 0.821                         & \textbf{0.825}       & 0.854                         & \textbf{0.862}       & 0.862                         & \textbf{0.864}       \\
FordB         & \textbf{0.824}                & 0.820                & \textbf{0.842}                & 0.836                & 0.843                         & \textbf{0.847}       \\
NIFECGT1 & \textbf{0.887}                & 0.855                & \textbf{0.906}                & 0.867                & \textbf{0.908}                & 0.860                \\
NIFECGT2 & \textbf{0.896}                & 0.856                & \textbf{0.912}                & 0.874                & \textbf{0.917}                & 0.879                \\
UWGLAll       & \textbf{0.640}                & 0.628                & 0.671                         & \textbf{0.682}       & 0.650                         & \textbf{0.677}       \\
Phoneme       & 0.405                         & \textbf{0.408}       & 0.414                         & \textbf{0.419}       & \textbf{0.442}                & 0.427                \\
Mallat        & \textbf{0.974}                & 0.971                & \textbf{0.976}                & 0.972                & \textbf{0.975}                & 0.969                \\
MSST          & 0.902                         & \textbf{0.914}       & \textbf{0.933}                & 0.931                & \textbf{0.940}                & 0.925                \\
MSRT          & 0.919                         & \textbf{0.922}       & 0.909                         & \textbf{0.933}       & 0.925                         & \textbf{0.942}       \\
SLC           & \textbf{0.947}                & 0.924                & \textbf{0.949}                & 0.936                & \textbf{0.944}                & 0.942                \\ \midrule
wins          & \textbf{6}                    & 4                    & \textbf{6}                    & 4                    & \textbf{6}                    & 4                    \\
p-value       & 0.232                         & -                    & 0.695                         & -                    & 0.492                         & -                    \\ \bottomrule
\end{tabular}}
\label{acc2}
\end{table}

\section{Conclusion}
In this paper, we study efficient semi-supervised time series classification with few labeled samples. We propose a new efficient graph-based time series classification, LB-SimTSC, with a new graph construction module that takes only linear time. Our proposed method resolves the scalability issue of constructing graphs from large datasets while maintaining the ability to allow time warping when comparing two time series. The experimental results on ten large datasets from the UCR archive show that our model achieves two orders of magnitude of speedup on graph construction without significant loss of classification accuracy compared to SimTSC. Furthermore, our work is the first to explore using a more efficient lower bounding-based graph to guide GCN while still maintaining the same ability as the more expensive DTW-based SimTSC.

\bibliography{main}

\newpage

\section{Appendix A: Experimental Details}
\subsection{Datasets} We provide detailed description of datasets in Table~\ref{datasets}. Ten selected datasets are all in large data volume, while having a wide range in the number of categories. They lie in 5 major time series data types: sensor, ECG, motion, simulation, and image. We also provide the abbreviation names of some datasets which we used in the main text due to space limitation. 

\subsection{Backbone Architecture}
We use the same backbone network as the default setting of SimTSC, which is a ResNet with three residual blocks. Each block is connected by shortcuts and consists of three 1D convolutional layers followed by batch normalization and a ReLU activation layer. The kernel size of three 1D convolutional layers are 7, 5 and 3, respectively. The number of channels for all convolutional layers are 64. Finally, a global average pooling is applied to the output of the last residual block. 

\subsection{Hyperparameters} 
We provide detailed hyperparameter settings of baselines and our model. 
\subsubsection{1NN-DTW} The warping window size is set to the minimum value between the length of time series and 100.
\subsubsection{SimTSC} We keep all the settings the same as original paper. That is, we set the scaling factor $\alpha$ to 0.3 and the number of neighbors $K$ to 3. The batch size is fixed to 128, and the number of epochs is fixed to 500. The number of GCN layer is set to 1. Adam with learning rate of $1\times 10^{-4}$ and weight decay of $4\times10^{-3}$ is used as the optimizer.  
\subsubsection{LB-SimTSC} We set 5\% of the length of time series as the warping range $r$ of LB\_Keogh, and use the same hyperparameters as SimTSC, except  $\alpha$. To find a suitable value of $\alpha$, we randomly select ten pairs of time series from the training set of all data and calculate their DTW and LB\_Keogh distances. We found that the distance calculated by LB\_Keogh is around 36 times larger than that of DTW. Thus, the distance scaling factor should also be around 36 times larger than 0.3. Therefore, we set $\alpha$ to be 11 ($0.3*36 \approx 11$).

\subsection{Accuracy Comparison}
\begin{figure}[htbp]
  \centering
  \includegraphics[scale=0.28]{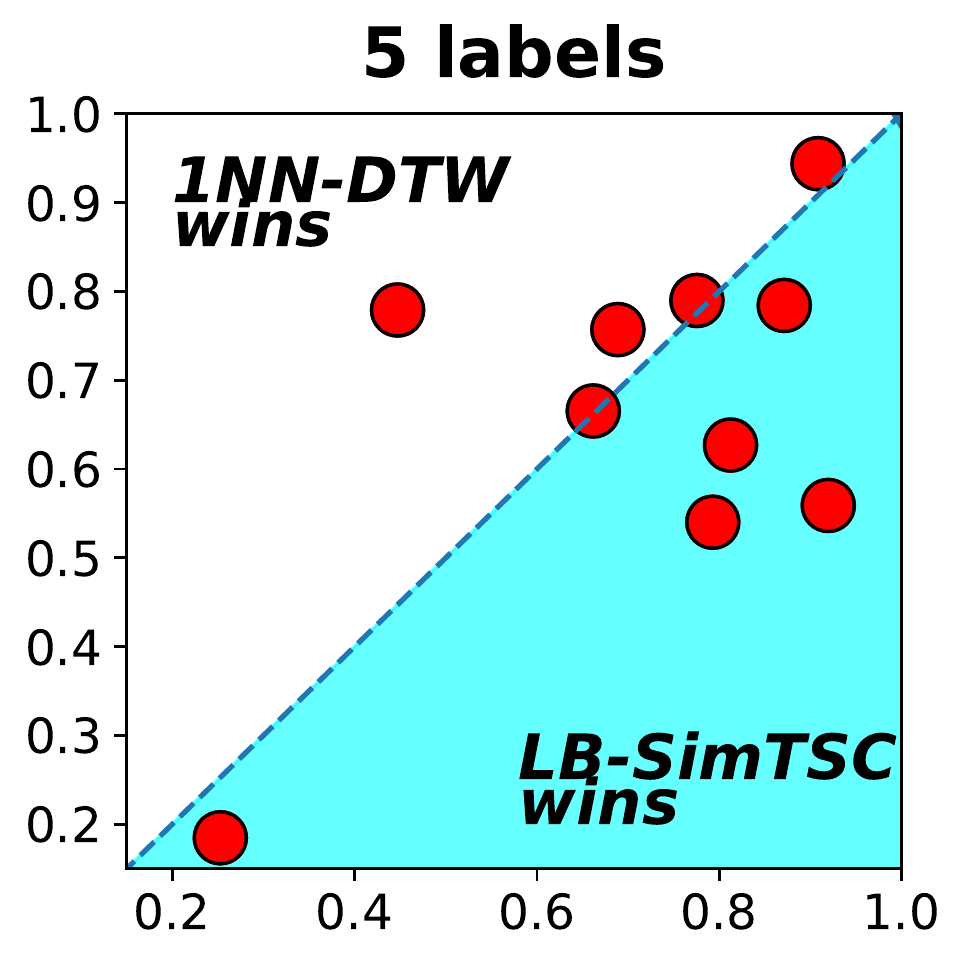}
  \includegraphics[scale=0.28]{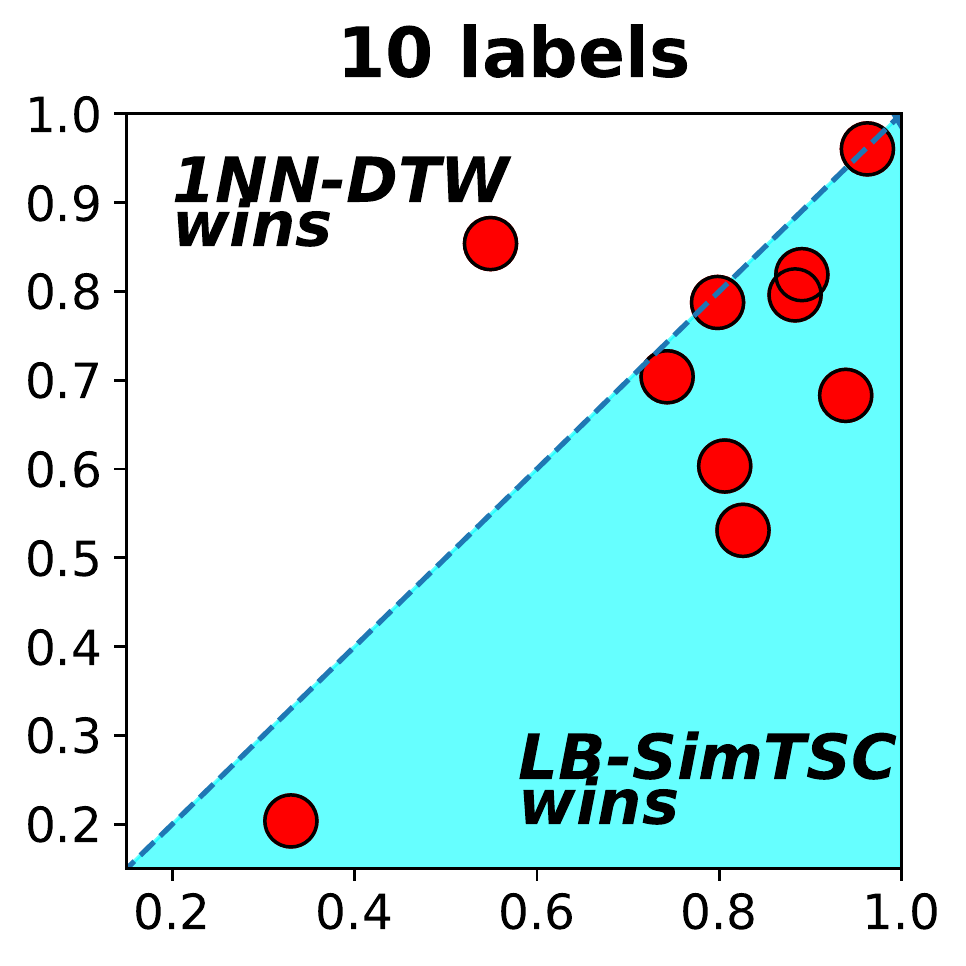}
  \includegraphics[scale=0.28]{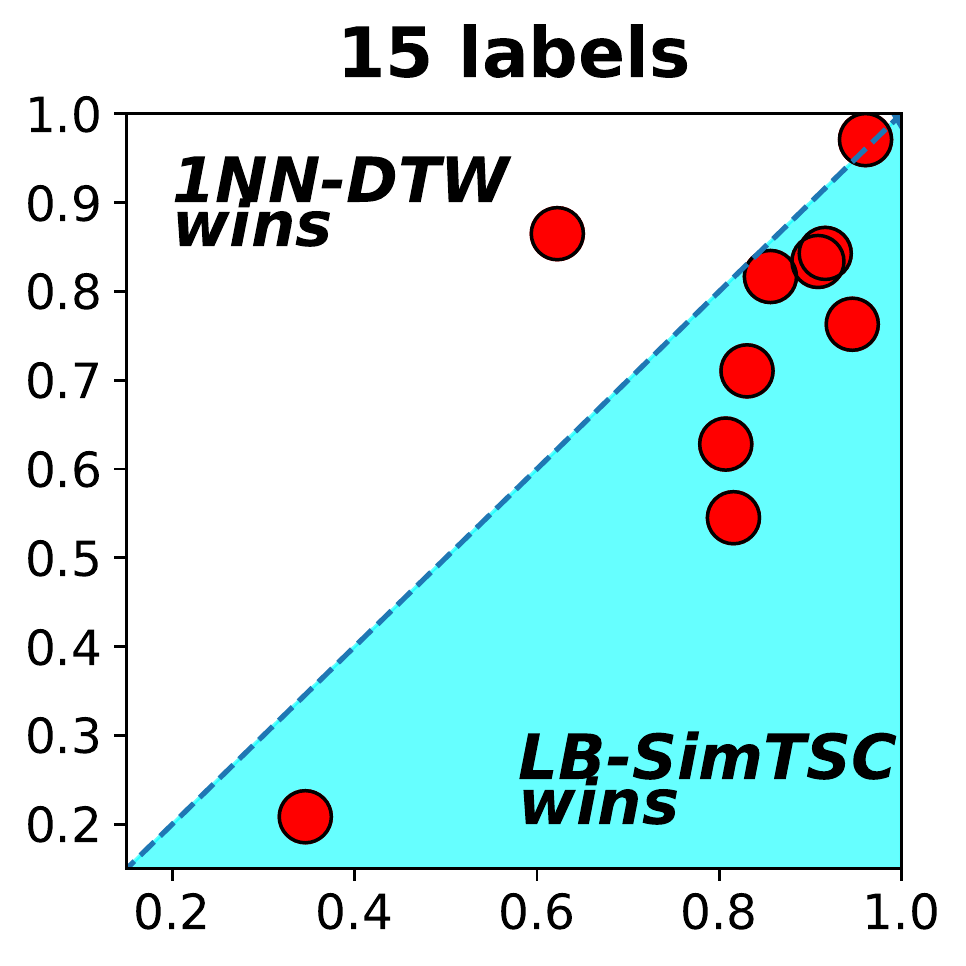}
  \includegraphics[scale=0.28]{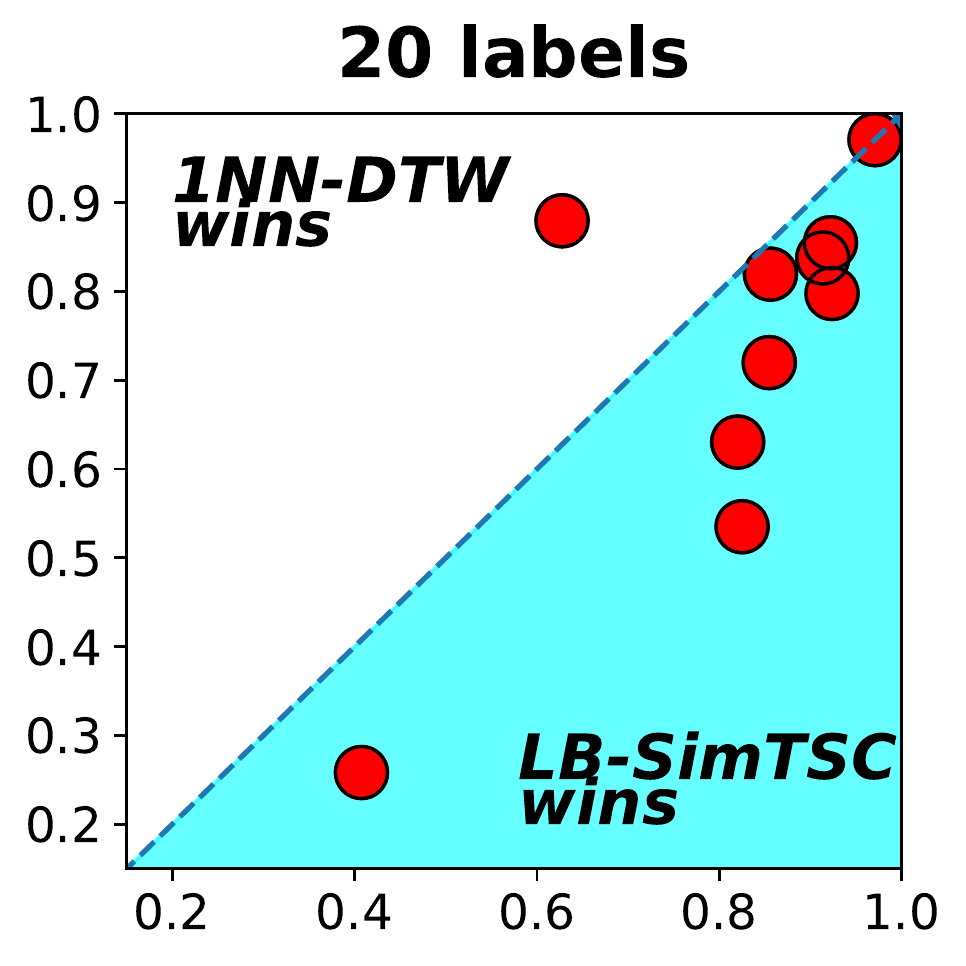}
  \includegraphics[scale=0.28]{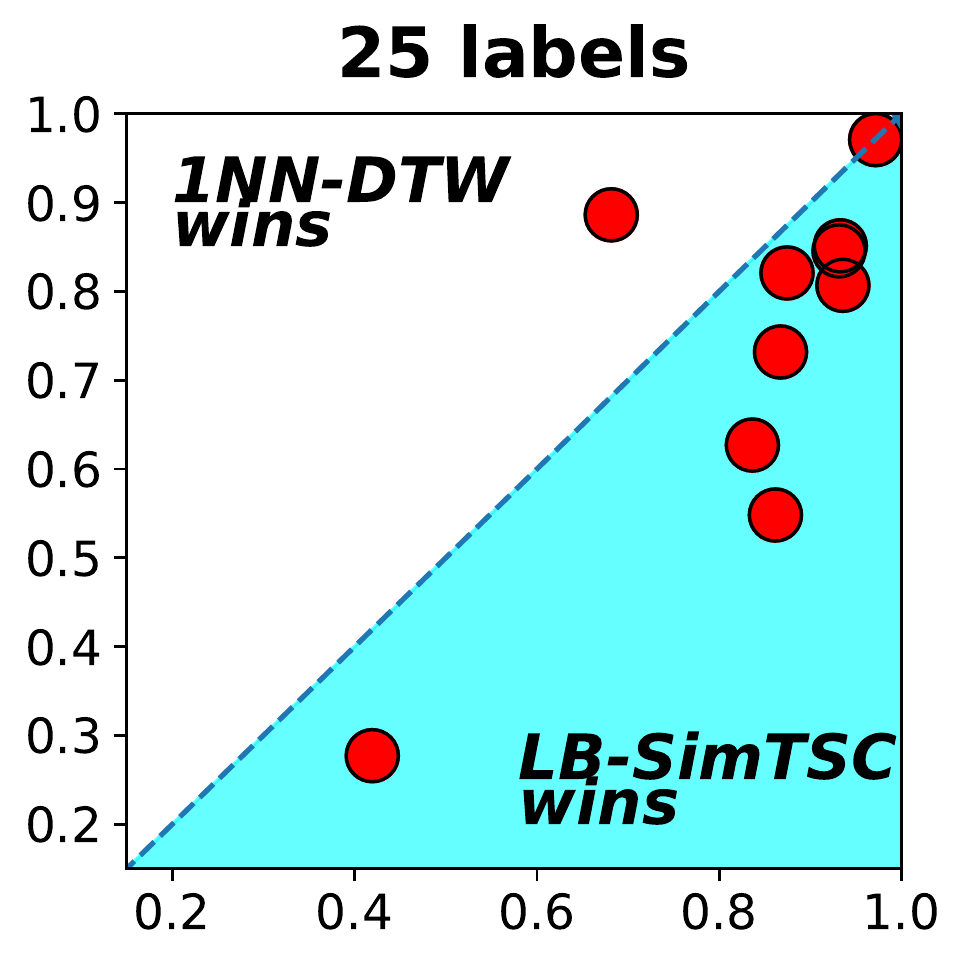}
  \includegraphics[scale=0.28]{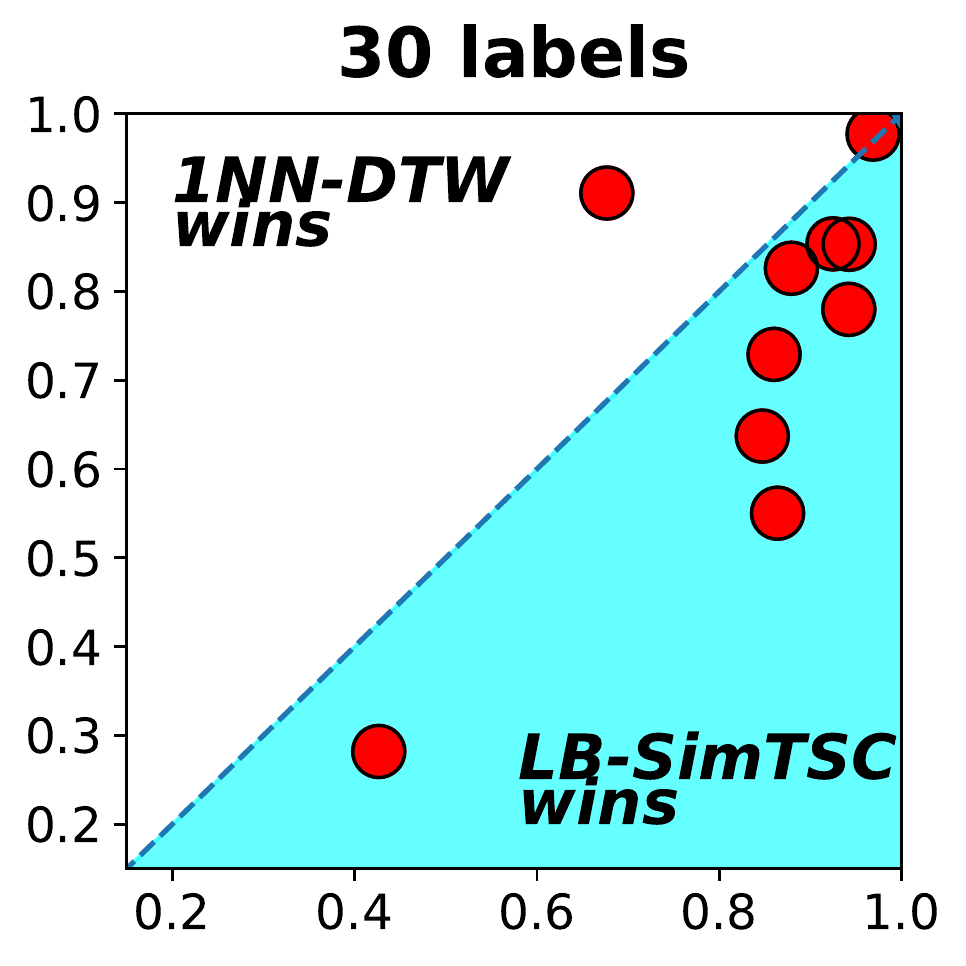}
  \caption{Classification accuracy: LB-SimTSC (right bottom shaded)  vs. 1NN-DTW (left top white). }
  \label{classification1}
\end{figure}

\begin{figure}[htbp]
  \centering
  \includegraphics[scale=0.28]{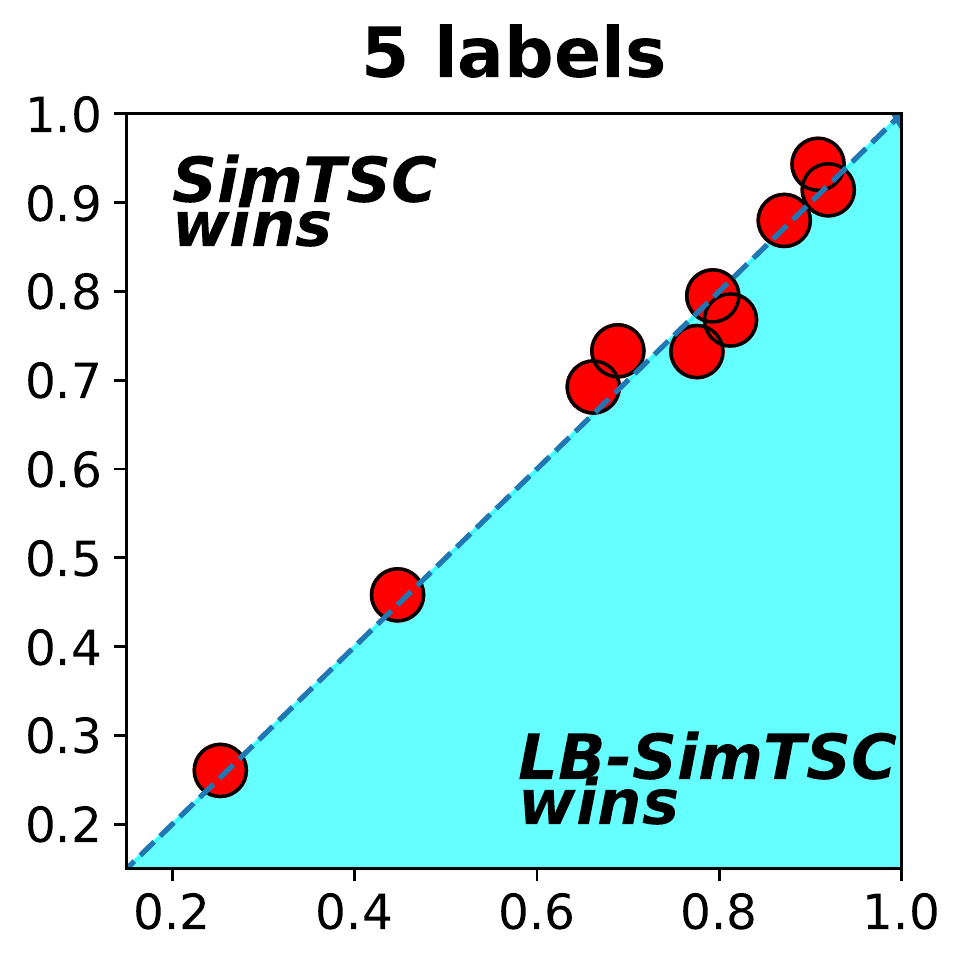}
  \includegraphics[scale=0.28]{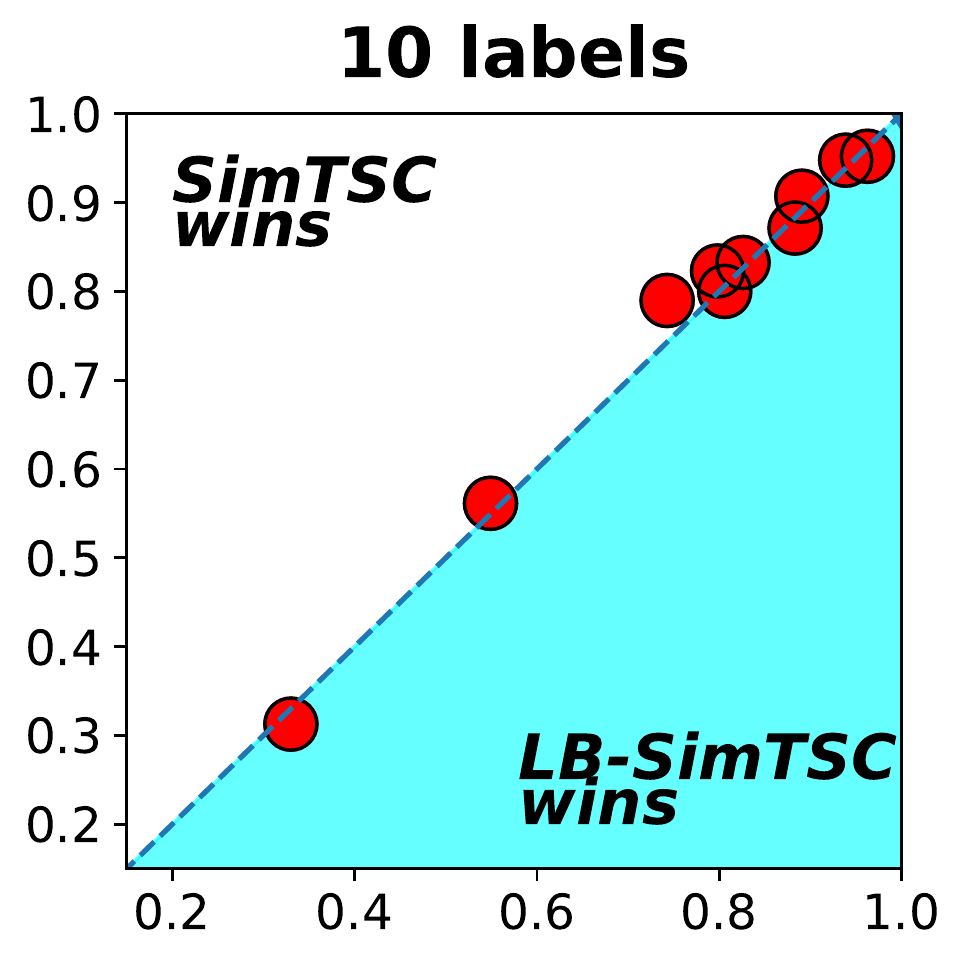}
  \includegraphics[scale=0.28]{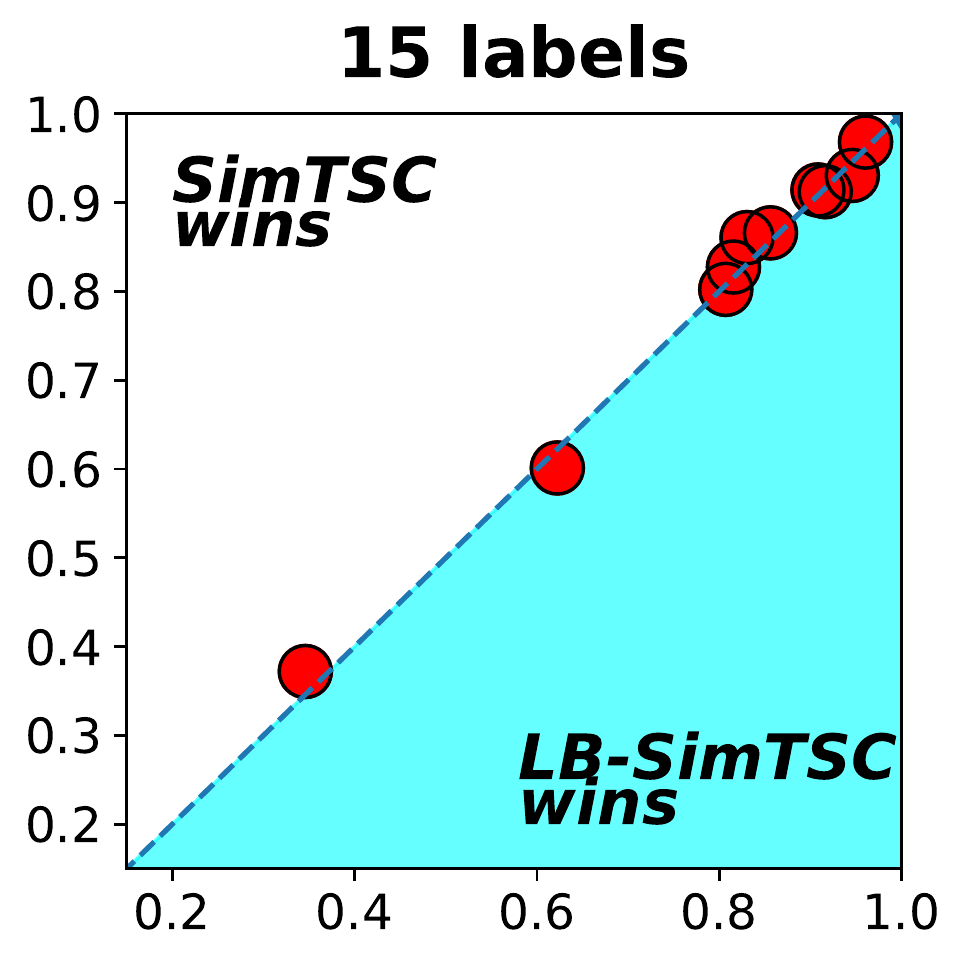}
  \includegraphics[scale=0.28]{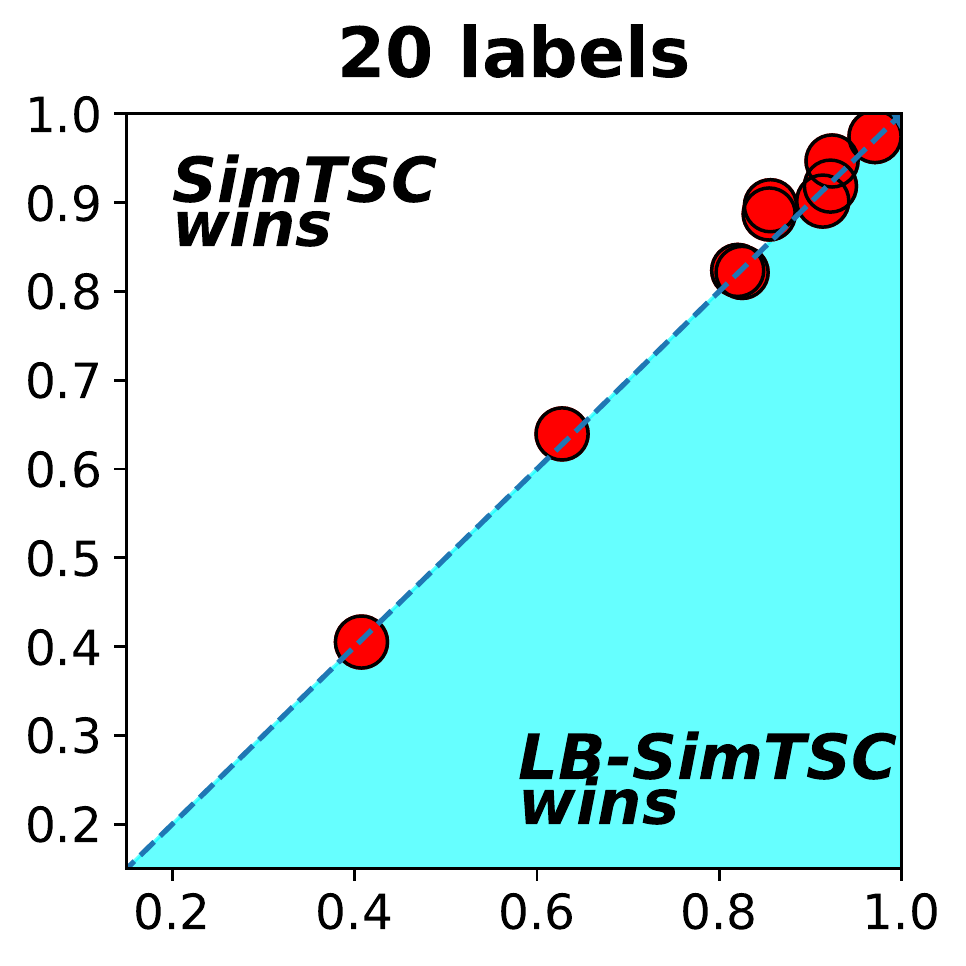}
  \includegraphics[scale=0.28]{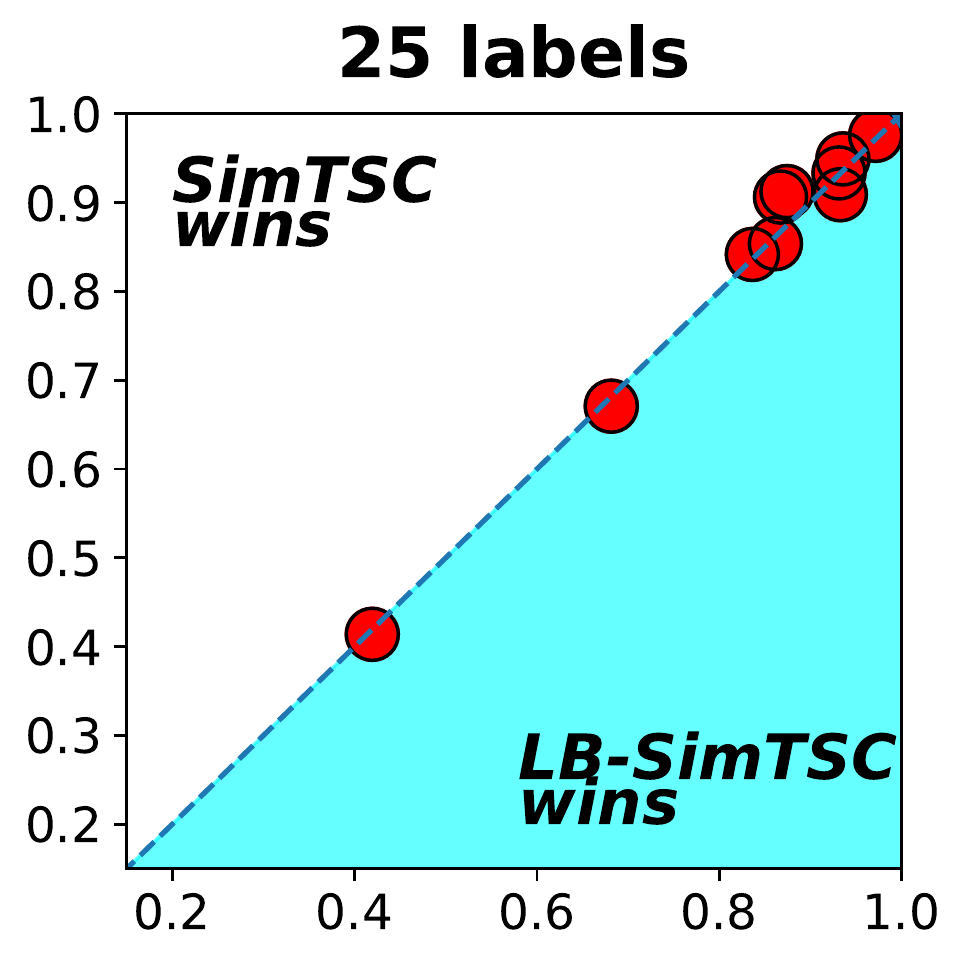}
  \includegraphics[scale=0.28]{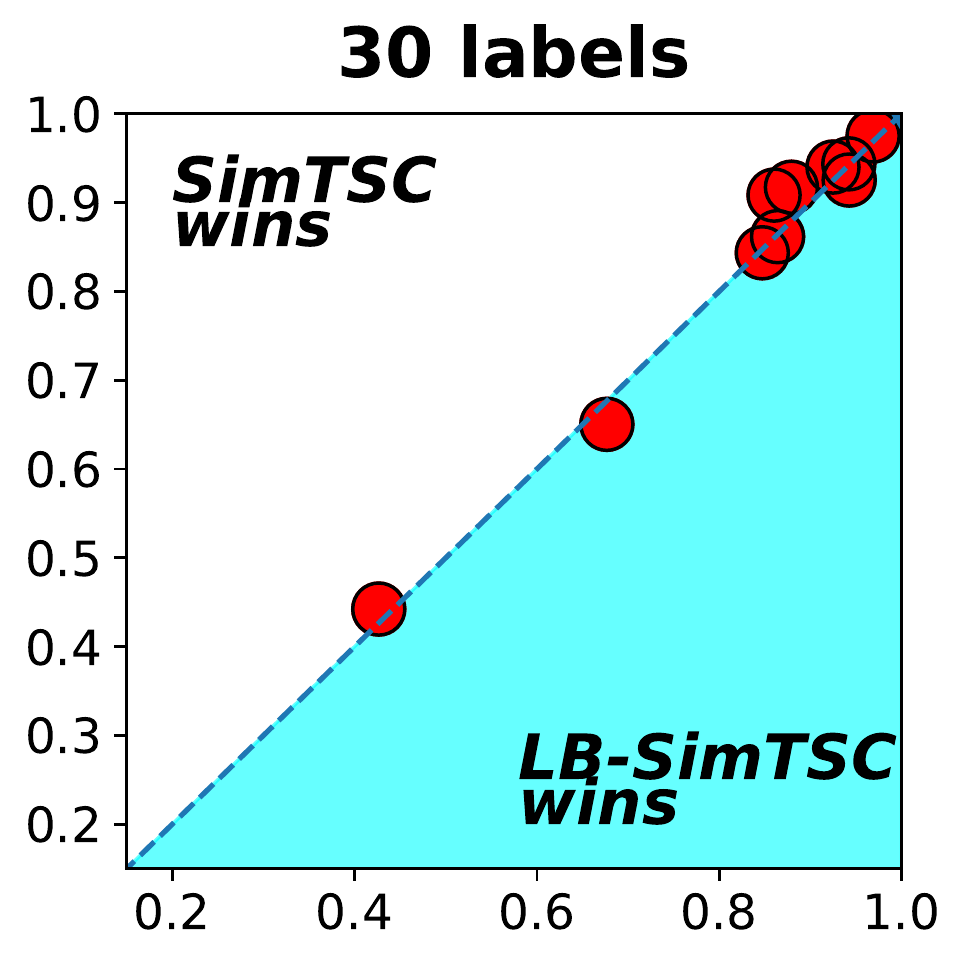}
  \caption{Classification accuracy: LB-SimTSC (right bottom shaded)  vs. SimTSC (left top white). }
  \label{classification2}
\end{figure}

Fig.~\ref{classification1} and Fig.~\ref{classification2} provide a visual summary of the results for accuracy. In Fig.~\ref{classification1}, the accuracy for our model is on the X-axis, and the accuracy for 1NN-DTW is on the Y-axis. Each point in a plot represents one dataset. If a point is in the lower triangle (shaded area), it means that our model wins. If a point is in the upper triangle (non-shaded area), it means that 1NN-DTW wins. The closer the point is to the midline, the smaller the gap between the two models. Except for 5 labels, our model clearly outperforms 1NN-DTW.
In Fig.~\ref{classification2}, the accuracy for our model is on the X-axis as well, and the accuracy for SimTSC is on the Y-axis. It can be observed that the accuracies of two models are very close. Therefore, we can conclude that two models do not have significant difference in accuracy.  

\begin{table}[htbp]
\caption{Description of ten large datasets}
\scalebox{0.7}{
\begin{tabular}{@{}llrrrr@{}}
\toprule
Datasets                   &Abbr.       & Type      & Class & Length & Instance \\ \midrule
FordA                      &-       & Sensor    & 2     & 500    & 4921     \\
FordB                      &-       & Sensor    & 2     & 500    & 4446     \\
NonInvasiveFetalECGThorax1 &NIFECGT1    & ECG       & 42    & 750    & 3765     \\
NonInvasiveFetalECGThorax2 &NIFECGT2    & ECG       & 42    & 750    & 3765     \\
UWaveGestureLibraryAll     &UWGLAll     & Motion    & 8     & 945    & 4478     \\
Phoneme                    &-     & Sensor    & 39    & 1024   & 2110     \\
Mallat                     &-      & Simulated & 8     & 1024   & 2400     \\
MixedShapesSmallTrain      &MSST        & Image     & 5     & 1024   & 2525     \\
MixedShapesRegularTrain    &MSRT        & Image     & 5     & 1024   & 2925     \\
StarLightCurves            &SLC         & Sensor    & 3     & 1024   & 9236     \\ \bottomrule
\end{tabular}}
\label{datasets}
\end{table}

\end{document}